\documentclass[a4paper,UKenglish,cleveref, autoref, thm-restate]{lipics-v2021}

\usepackage{booktabs}
\usepackage{multirow}

\title{Georeferencing Non-Gazetteered Place Names using Biological Specimen Records}

\author{Aneesha Fernando}{School of Mathematical and Computational Sciences, Massey University, Auckland, New Zealand}{aneeshafedo@gmail.com}{https://orcid.org/0009-0000-8088-6646}{}
\author{Surangika Ranathunga}{School of Mathematical and Computational Sciences, Massey University, Auckland, New Zealand}{S.Ranathunga@massey.ac.nz}{https://orcid.org/0000-0003-0701-0204}{}

\author{Kristin Stock}{School of Mathematical and Computational Sciences, Massey University, Auckland, New Zealand}{K.Stock@massey.ac.nz}{https://orcid.org/0000-0002-5828-6430}{}

\author{Raj Prasanna}{Joint Centre for Disaster Research, Massey University, Wellington, New Zealand}{R.Prasanna@massey.ac.nz}{https://orcid.org/0000-0002-5608-6558}{}

\author{Christopher B. Jones}{School of Computer Science and Informatics, Cardiff University, United Kingdom}{jonescb2@cardiff.ac.uk}{https://orcid.org/0000-0001-6847-7575}{}

\authorrunning{A. Fernando, K. Stock, S. Ranathunga, R. Prasanna, and C. B. Jones}

\Copyright{Aneesha Fernando, Surangika Ranathunga, Kristin Stock, Raj Prasanna, and Christopher B. Jones} 



\supplementdetails{Dataset}{https://doi.org/10.6084/m9.figshare.32397048}

\funding{This work was supported by the Ministry of Business Innovation and Employment Smart Ideas Fund (grant number MAUX2104), New Zealand.}

\nolinenumbers 

\EventEditors{Sabine Timpf, Gabriele Filomena, Armand Kapaj, Rui Zhu, Nicholas Giudice, and Ed Manley}
\EventNoEds{6}
\EventLongTitle{17th International Conference on Spatial Information Theory (COSIT 2026)}
\EventShortTitle{COSIT 2026}
\EventAcronym{COSIT}
\EventYear{2026}
\EventDate{September 22--25, 2026}
\EventLocation{York, UK}
\EventLogo{}
\SeriesVolume{393}
\ArticleNo{8}

\begin{document}

\maketitle

\begin{abstract}
Biological specimen records collected by natural history institutions constitute a rich source of temporal geographic knowledge, capturing biodiversity information about regional landscapes as they were recorded at different times. Using digitised data from the Allan Herbarium (New Zealand), this study identifies place names in these specimen locality descriptions that are absent from current gazetteers; we refer to these as non-gazetteer place names (NGPs). These place names are typically historical, vernacular, or colloquial and were used as landmarks to describe a specimen's location at the time of collection. We then investigate the problem of georeferencing the NGPs using only the limited information available in the specimen records. To resolve this, we leverage repeated occurrences of the same place name across specimen records with different specimen locations and spatial relation terms, extracting and inverting these relations to derive constraints on NGP locations. This approach is instantiated within deterministic, probabilistic, and LLM-based methods, enabling a comparative analysis of their strengths and limitations for text-based spatial inference. On a pseudo-NGP benchmark, probabilistic inference achieves the highest accuracy (median error 1.43 km; A@1 km 36\%), while the LLM yields competitive but less precise estimates (median error 1.80 km; A@1 km 31\%), indicating that, despite advances in LLMs, traditional modelling remains advantageous when high spatial precision is required.

\end{abstract}

\section{Introduction}
\label{sec:introduction}
The content and metadata associated with biological specimen collection records constitute a rich and underutilised source of geographic information. These records usually include locality descriptions that are routinely recorded digitally by natural history institutions, including museums and herbaria \cite{chapman_georeferencing_2020, wieczorek_point-radius_2004}. A typical locality description specifies the location where a biological specimen was collected and usually contains at least one place name, often several, along with spatial relation terms that describe the specimen’s position relative to those places. Such records span more than a century and capture geographic information as it was recorded at the time of collection. Consequently, some of the place names recorded in these descriptions are not represented in current gazetteers and can be classified as non-gazetteer place names (NGPs). We define NGPs as place names that are entirely absent from existing gazetteers, including alternative names and other variants, and therefore have no associated spatial footprint in those gazetteers. Their absence may stem from temporal changes in toponymy, the use of informal or locally recognised names, or incomplete coverage of certain regions and feature types in existing gazetteers \cite{hill2009georeferencing}. Given that millions of such records are available worldwide, these descriptions represent a significant and largely untapped source of geographic knowledge. Incorporating these place names, together with their inferred spatial footprints, into digital gazetteers has the potential to improve spatial reference coverage and support more robust place name resolution, not only for specimen georeferencing but also for broader geographical information retrieval applications and analyses.

In this research, we focus on georeferencing NGPs using information available within biological specimen records. The information contained in biological specimen records includes free-text locality descriptions indicating where specimens were collected, coordinates of collection locations, habitat descriptions, and higher-level administrative units such as states or provinces.\footnote{Example Allan Herbarium specimen record: \url{https://scd.landcareresearch.co.nz/Specimen/CHR\%20508209}.} These elements are used jointly for both georeferencing and disambiguation of place names in locality descriptions, with specimen coordinates serving as important spatial anchors that constrain the interpretation of relative spatial relations and reduce uncertainty in the inferred locations of unknown place names.

Biological specimen collection practices further provide a source of spatial redundancy that can prove valuable. During fieldwork, collectors typically traverse a specific route and record multiple specimens collected within a limited geographic area. As a result, many specimens share common geographic features and place names across their locality descriptions. In modern collection workflows, specimen coordinates are often recorded in the field and subsequently validated by georeferencers. In historical datasets, however, coordinates may also have been assigned retrospectively from locality descriptions, and as a result, some uncertainty may be present in the coordinates. Moreover, a substantial proportion of records, particularly those collected prior to the widespread adoption of GPS, remain ungeoreferenced, and existing georeferences vary in quality \cite{marcer2021quality}. To address this gap, a range of georeferencing methods has been developed, from time-intensive manual interpretation using GIS tools and maps \cite{wieczorek_point-radius_2004} to automated approaches based on rule-based techniques \cite{van_erp_georeferencing_2015}, machine learning \cite{scott_automated_2021}, and modern Large Language Models (LLMs) \cite{Fernando21012026}. Building on this context, our methodology leverages digitised specimen records and exploits the repeated occurrence of the same NPG across multiple locality descriptions as a source of spatial evidence. When an unknown place name appears in several descriptions, the spatial constraints expressed in each record are combined to infer its most plausible location.

Table \ref{tab:cabstand-example} illustrates this pattern of repeated place name occurrence across multiple locality descriptions. The NGP \emph{Cabstand}\footnote{The publication "Banks Peninsula Conservation Walks" \cite{doc_banks_peninsula_walks}, issued by the Department of Conservation, New Zealand, refers to “Cabstand” as the name of a junction on Summit Road in Banks Peninsula.} appears in four independent locality descriptions, each linked to a distinct specimen and to the coordinates of the corresponding specimen location. 
Each description employs relative spatial relation terms such as “north of,” “south of,” or “southeast of” to describe the specimen’s location relative to “Cabstand,” thereby imposing qualitative constraints on its location. Taken together, these multiple references impose complementary spatial constraints that can be exploited to infer the location of the NGP by inverting (reversing) the spatial relations. The associated specimen coordinates act as spatial anchors, bounding the plausible region in which Cabstand must lie. By integrating reversed relative direction terms with the spatial distribution of specimen locations, it becomes possible to approximate the location of "Cabstand" even in the absence of an explicit gazetteer entry. 

To operationalise this idea, we investigate three complementary georeferencing methodologies. The first approach is a deterministic method that draws on the formal search space model of Chen et al.\ \cite{chen_georeferencing_2018} and constructs an Approximate Location Region (ALR) for an NGP by modelling the spatial relationship constraints expressed in locality descriptions relative to specimen coordinates. When the same place name appears across multiple descriptions, the intersection of their respective constraint regions progressively reduces the ALR, yielding increasingly precise candidate locations. Second, we adopt a probabilistic framework in which spatial relations are represented as likelihood functions conditioned on specimen coordinates and referenced place names. Multiple occurrences are aggregated to estimate the most likely location from the resulting spatial probability distribution, replacing crisp geometric boundaries with a continuous representation. Third, we explore an LLM-based approach by providing locality descriptions and specimen coordinates as inputs in the prompt and tasking the LLM with inferring NGP locations from combined textual and numeric cues. This allows us to assess the extent to which LLMs can integrate spatial language and coordinate information for georeferencing.

Among these three approaches, the probabilistic method yields the most precise results, although the other approaches also perform competitively. To our knowledge, this study represents the first systematic investigation of using biological specimen records as a primary data source for extracting and georeferencing NGPs. The methodologies and findings presented here demonstrate the feasibility and value of exploiting specimen-derived spatial information for enriching digital gazetteers.

The contributions of this work are as follows:
\begin{bracketenumerate}
\item We demonstrate that biological specimen records contain non-gazetteered place names, highlighting these records as an underutilised source of geographic information over time. 
\item We propose a data-driven strategy to georeference non-gazetteered place names using biological specimen records by exploiting multiple occurrences of the same place name across different specimen locality descriptions to derive spatial constraints on unknown locations.
\item We provide a systematic comparison of deterministic, probabilistic, and LLM-based approaches to text-based spatial modelling and inference of the locations of non-gazetteered place names, evaluating their strengths and limitations.
\item We construct a new benchmark dataset for evaluating methods for georeferencing place names derived from biological specimen records, which contain rich spatial knowledge.
\end{bracketenumerate}

The remainder of this paper is organised as follows. Section~2 reviews related work, Section~3 details the methodology. Section~4 presents the results, and Section~5 concludes the paper with a discussion of limitations and future research directions.

\begin{table}
\caption{Example illustrating how multiple specimen locality descriptions refer to the same non-gazetteered place name (“Cabstand”), each accompanied by distinct coordinates and expressed through varying spatial relation terms (e.g., near, north of, south of, southeast of) that describe the NGP’s position relative to the specimen location.}
\label{tab:cabstand-example}
\centering
\begin{tabularx}{\textwidth}{lXll}
\toprule
\textbf{Place name} & \textbf{Locality description} & \textbf{Latitude} & \textbf{Longitude} \\
\midrule
\multirow{4}{*}{Cabstand}
 & Banks Peninsula, Brocheries Road near Cabstand
 & -43.8030 & 173.0210 \\
 & Banks Peninsula, northeast of Akaroa, north of Cabstand, Summit Road
 & -43.7949 & 173.0210 \\
 & Banks Peninsula, south of Cabstand
 & -43.8093 & 173.0172 \\
 & Banks Peninsula, Long Bay Road southeast of Cabstand
 & -43.8039 & 173.0259 \\
\bottomrule
\end{tabularx}
\end{table}

\section{Related Work}
\label{sec:related-work}

This section reviews research relevant to the georeferencing of place names absent from gazetteers. It first outlines general approaches to georeferencing place names from textual descriptions, which provide the foundation for the research problem, and then examines studies addressing non-gazetteered place names, clarifying how this problem differs from generic georeferencing and highlighting the existing gap in the literature.

Automated georeferencing links textual descriptions or other location-referencing information to geographic coordinates. While some methods focus on georeferencing entire textual documents, such as Wikipedia articles, news reports, or social media posts \cite{wikigeorefOlivier, han2014text}, others focus specifically on georeferencing the place names mentioned within text descriptions. The latter typically involves two main steps: identifying place names (toponym recognition) and determining their correct locations (toponym resolution or geocoding) \cite{liu_geoparsing_2022, wang_are_2019}. In many applications, both steps are closely tied to gazetteers: recognition benefits from name lists, and resolution selects among candidate gazetteer entries.

Early georeferencing approaches treated toponyms as gazetteer lookups with heuristic disambiguation, favouring populous or administratively prominent places \cite{alex2019geoparsing, alex2016homing}. Unsupervised models improved this by selecting spatially consistent candidate sets using only gazetteer data \cite{Kamalloo2018A}. However, a fundamental limitation of gazetteer-dependent methods is their incomplete coverage of vernacular, colloquial, and fine-grained place names \cite{hill2009georeferencing, acheson2017quantitative, smart2010multi}. As Wang et al. \cite{wang_are_2019} observe, many geoparsing corpora restrict annotations to city-level and above, excluding fine-grained toponyms such as street names or the names of parks and monuments. These methods also struggle to resolve historical or outdated names that are no longer present in modern geographic databases \cite{gritta2018s}. Nevertheless, such names frequently persist in literary, archival, and other domain-specific textual sources with a historical dimension \cite{alex2019geoparsing}. To overcome this, gazetteer-independent approaches such as TopoCluster learn spatial word distributions from large georeferenced corpora (GeoWiki) and resolve toponyms based on contextual geographic overlap \cite{DeLozier_Baldridge_London_2015}.

Deep learning approaches replaced hand-crafted heuristics with learned contextual representations. For toponym recognition, neural sequence models (e.g., BiLSTM-CRF and CNN–RNN hybrids) capture informal spellings and noisy text, outperforming rule-based and gazetteer-only systems \cite{Aldana-Bobadilla2020Adaptive, Zhou2023TopoBERT:, 2022Chinese}. For toponym resolution, neural models either predict continuous coordinates from contextual embeddings, treating the earth as a regression surface or frame the task as tile classification or shared embedding learning, enabling similarity-based grounding beyond strict gazetteer lookup \cite{Cardoso2021A, Fize2021Deep}. However, these methods rely on large-scale training corpora and still presuppose that targets resemble places represented in their training distributions.

LLMs are beginning to reshape both toponym recognition and resolution by injecting stronger, more explicit geographic knowledge into language representations. Geospatially grounded transformers such as GeoLM \cite{Li2023GeoLM:} and GeoReasoner \cite{Yan2024GeoReasoner:} augment standard pretraining with coordinate‑based spatial embeddings and contrastive alignment between textual and geospatial contexts, then fine‑tune (or zero‑shot apply) these models to sequence‑tagging for toponym recognition and toponym linking against gazetteers, achieving state‑of‑the‑art or near state‑of‑the‑art performance on multiple benchmarks. For large‑scale resolution, lightweight open‑source LLMs (e.g., fine‑tuned Mistral, Llama‑2, Baichuan2) can be trained to infer an unambiguous type‑level referent (city/state/country) from context and then delegate to geocoders such as GeoNames or Nominatim, substantially outperforming previous neural and voting‑based resolvers at the global scale \cite{Hu2024Toponym}. 

In our study, we target a stricter case of place names that are not found in gazetteers at all, not even as alternate spellings, variants, or official forms. In this setting, resolution cannot fall back on candidate resolution from GeoNames, OpenStreetMap, or similar resources and must infer location from other evidence. To date, relatively few studies have directly targeted the resolution of such truly non-gazetteered place names. A closely related line of work is the Bhugol framework proposed by Sharma et al.~\cite{sharma2023spatially}, which addresses non-gazetteered place names by exploiting external regularities rather than relying on an explicit gazetteer. Bhugol estimates coordinates by clustering either gazetteer place names sharing frequent morphological affixes (via HDBSCAN) or gazetteer place names that co-occur with the target name in a news corpus (via DBSCAN), and assigning the centroid of the most representative spatial cluster as the inferred location; an integrated variant combines both signals. Among existing approaches, this methodology is the closest to our work in its explicit focus on non-gazetteered place names. However, it is strongly region-dependent and relies on recurring naming conventions and extensive corpus evidence, which limits its portability across geographies and languages. Moreover, even its best-performing variant, Bhugol-LS, reports mean distance errors often exceeding 100 km, which may be acceptable for coarse-resolution tasks but falls short in applications requiring fine-grained spatial accuracy.

Chen et al.~\cite{chen_georeferencing_2018} propose a graph-based approach to georeferencing place names absent from gazetteers by exploiting spatial relationships extracted from natural language, a method that has informed our work. Their framework uses place graphs from collective manually created descriptions, with nodes representing locations and edges encoding qualitative spatial relation terms, enabling inference of a target location by linking a non-gazetteered place name (locatum) to surrounding gazetteered landmarks (relata). However, their approach assumes that the relatum can be reliably identified and disambiguated, and that the locatum corresponds to a formal or alternative gazetteer entry. In contrast, the locality descriptions in our data refer to  locations that lack any representation, formal or variant in existing gazetteers or VGI resources. Moreover, our records provide no external contextual cues regarding the prominence or granularity of either locatum or relatum, making it infeasible to construct the probabilistic search space required by their pipeline. Relatedly Yousaf and Wolter ~\cite{yousaf2022reasoning} propose a reasoning-based model that interprets natural language place descriptions by representing named and unnamed spatial entities as relational constraints and resolving them through spatio-ontological inference. While their approach aligns with our emphasis on spatial relation terms, it ultimately grounds interpretations by matching inferred entities to OpenStreetMap features, whereas our work targets place names that lack any gazetteer representation.

Taken together, the literature suggests a gap in both (i) using specimen records as a source for identifying place names that are absent from gazetteers and (ii) developing methods to georeference place names that are neither gazetteered nor supported by rich corpus-derived evidence. This motivates our effort to extract and georeference non-gazetteered place names using biological specimen records as the source.

\section{Methodology}
\label{sec:methodology}

\subsection{Locality description corpus}
\label{sec:data-prep}
For our experiments, we use biological specimen records from the Allan Herbarium (CHR)\footnote{https://www.landcareresearch.co.nz/tools-and-resources/collections/allan-herbarium} dataset of Manaaki Whenua – Landcare Research (MWLR), New Zealand. We retain only records that include both specimen coordinates and a free-text locality description. As described earlier, specimen datasets often contain repeated locality descriptions as multiple specimens may be collected at the same site. Because our objective is to use distinct locality descriptions to constrain an NGP's position, we removed duplicate locality descriptions from the original MWLR extract. We also noted the presence of records that share identical coordinate pairs but differ slightly in wording (e.g., \emph{``Campbell Island, near \textbf{footpath towards} Met Service \textbf{hostel}''} vs.\ \emph{``Campbell Island, near \textbf{concrete pad on} Met Service \textbf{base}''}). To avoid redundant spatial evidence, we removed records with duplicate coordinate pairs, keeping only the first. While this filtering reduces redundancy, it may also discard useful linguistic variation, as records with the same coordinates can contain complementary contextual clues.

Next, we extracted place names from locality descriptions using the \emph{en\_core\_web\_trf} Named Entity Recognition (NER) model in spaCy\footnote{https://spacy.io/}, fine-tuned on a subset of 50 manually annotated locality descriptions from our data. We then retained only records in which at least one extracted place name participates in an explicit spatial relationship (e.g., ``18 km west of Ashburton''). Descriptions that merely list place names without explicitly expressing a spatial relation (e.g., ``Green Hill, Upper Poulter Valley, Waimakariri Region'') were excluded, although in some cases such listings may implicitly encode hierarchical relationships.

The initial list of relation terms was developed from common spatial expressions observed during manual inspection of locality descriptions and was subsequently refined iteratively by checking unmatched and incorrectly matched examples. The matching patterns followed a general structure of the form \texttt{<spatial relation term> <place name>}. When a locality description contained multiple such expressions, all matching relations were extracted. A detailed discussion of the spatial relation terms identified in the dataset is provided in Section~\ref{sec:interpreting-spatial-relations}.
  
After processing, the dataset comprised 13,746 locality descriptions, grouped by extracted place names, together with their associated spatial relations and the coordinates of each locality description.

\subsection{Identifying non-gazetteered place names}

To demonstrate the presence of NGPs in biological specimen records, we used a two-stage identification procedure: (i) automated screening to construct a candidate set, and (ii) targeted manual verification to assess its reliability. This procedure was applied to the place names extracted from the above-mentioned dataset of 13,746 locality descriptions. Each extracted place name was checked against the New Zealand Geographic Board (NZGB)\footnote{https://gazetteer.linz.govt.nz} Gazeteer, GeoNames\footnote{https://www.geonames.org}, and OpenStreetMap (OSM)\footnote{https://www.openstreetmap.org} to determine whether it is already gazetteered.

In the automated stage, we applied sequential filtering using these three gazetteers. First, each candidate was compared against the downloaded NZGB Gazetteer export and matching names were removed. Second, the remaining candidates were cross-checked against the GeoNames export for New Zealand to identify additional gazetteered names not present in the NZGB dataset. Third, unresolved candidates were queried against the OSM Nominatim API, and any strings that could be matched to an OSM feature were excluded. After these steps, 520 candidates were not found in any of the three gazetteers and were retained as candidate NGPs. To reduce sensitivity to minor orthographic variation in the free-text locality field (e.g., misspellings, spacing differences, or abbreviations), we employed fuzzy string matching rather than exact matching, using a similarity threshold of 80\% with the RapidFuzz library\footnote{https://pypi.org/project/RapidFuzz}.

Despite this multi-source filtering, the resulting set still required manual verification because specimen locality descriptions are authored as free text and are not standardised. To quantify the extent of these issues, we manually inspected a random sample of 50 of the 520 automatically identified candidates. Of these, 30 were confirmed as true NGPs.

The observed false positives largely fell into five categories:

\begin{itemize}
    \item \textbf{Alternative or informal naming} (a real place, but referenced using a non-standard variant): e.g., Bowscale Lake for Bowscale Tarn; Hutt Road for Hutt Motorway.
    \item \textbf{Generic feature terms misidentified as place names by spaCy} (common nouns treated as named entities): e.g., Airport in “Kaitaia, Quarry Road, near Airport”; BNZ bank in “Waikanae, marae grounds behind BNZ bank”.
    \item \textbf{Misspellings} (orthographic noise that persisted despite fuzzy matching): e.g., Mt Rockfort for Mt Rochfort; Baffalo beach for Buffalo Beach.
    \item \textbf{Abbreviations} (shortened forms not captured by gazetteer matching): e.g., big r. for Big River; Cobb V. for Cobb Valley.
    \item \textbf{Compound or delimited forms} (multi-name strings that complicate recognition and matching): e.g., "along the Slaughter/Long Burn"; "near Belltown/Manunui hut".
\end{itemize}

These error modes highlight that identifying truly non-gazetteered names is not simply a matter of gazetteer lookup. Developing a comprehensive recognition and normalisation pipeline is an important problem in its own right; however, our aim here is to establish that specimen records themselves are a credible source of non-gazetteered names, and the findings in our analysis support this.

Based on the manual audit, we estimated that approximately 60\% of the automatically extracted candidates were true NGPs. Applying this precision estimate to the 520 candidates yields an expected $\sim 312$ true NGP mentions, corresponding to roughly 2\% of the 13{,}746 locality descriptions in our sample (i.e., $520 \times 0.60 \approx 312$; $312/13{,}746 \approx 0.023$). Although this proportion appears small, it is operationally significant at the collection scale: biodiversity infrastructures aggregate millions of specimen records, and even a 2\% rate translates into tens of thousands of locality descriptions containing place references that cannot be resolved through conventional gazetteer-based workflows. Notably, this proportion is observed in New Zealand, a relatively well-mapped country with a mature national gazetteer and rich VGI sources, suggesting that the prevalence of NGPs is likely to be higher in regions with less complete mapping coverage. 

Table~\ref{tab:example_ngps} presents a representative subset of NGPs from our dataset, categorised according to the most commonly observed (non-exhaustive) NGP types. Many of these NGPs correspond to historical place names that are absent from current gazetteers, yet retain substantial historical value.

\begin{table}[!b]
\centering
\caption{Examples of manually verified NGPs identified in the dataset, categorised by the most commonly observed (non-exhaustive) types, together with the locality descriptions in which they appear and the corresponding record dates.}
\label{tab:example_ngps}
\begin{tabularx}{\textwidth}{p{2cm} p{3.5cm} X p{2cm}}
\toprule
\textbf{Type} & \textbf{Place name} & \textbf{Example occurring locality description} & \textbf{Recorded date} \\
\midrule
Farm stations & Beltana station & North Canterbury, hills east of Parnassus, above Beltana Station, near radio station & 6/01/1969 \\
& Ngatapa Sunworth Station & Gisborne, northwest of Ngatapa Sunworth Station. & 24/02/1990 \\
\midrule
Trig stations & Oropuke trig & Chatham Island, nearest major locality Oropuke trig, Southern Tableland, east of Oropuke trig, at top end of ``The Slump'', near hut & 26/02/1985 \\
& Driscoll Trig & Wairarapa, East Wairarapa Ecological District, Te Wharau near Driscoll Trig & 7/08/1992 \\
\midrule
Infrastructure facilities & Huia Filter Station & Titirangi, Woodlands Road, at Huia Filter Station & 16/03/1995 \\
& P.\ \& T.\ Microwave Station & Vernon Hills, Marlborough, near P.\ \& T.\ microwave station & 24/11/1970 \\
\midrule
Community facilities & Bryant Home & South shore of Raglan Harbour, near Bryant Home & 23/10/1967 \\
& Manaroa School & Forest remnant beside Manaroa school near shore, Pelorus Sound, Marlborough & 12/11/1971 \\
\midrule
Recreation places & Christian Youth Camp & Whakamaru, near Christian Youth Camp & 25/02/1983 \\
& Catleys (camp site) & Kauaeranga Valley, Thames opposite Catleys & 26/11/1973 \\
\midrule
Hospitality venues & Cave Rock Tearooms & Sumner near Christchurch, Canterbury, near Cave Rock Tearooms & 28/10/1971 \\
& Delta Lodge & Near Delta Lodge, Orongorongo Vly. & 25/06/1974 \\
\bottomrule
\end{tabularx}
\end{table}

\subsection{Georeferencing non-gazetteered place names}
We present three methods for georeferencing NGPs using biological specimen records: deterministic, probabilistic, and LLM-based. Each method has distinct strengths and weaknesses. The deterministic and probabilistic methods use structured relation inputs derived from locality descriptions, whereas the LLM-based method operates directly on the unprocessed locality descriptions and associated specimen coordinates. However, as noted in the Introduction, all three are based on the same underlying strategy of exploiting repeated occurrences of the same NGP across multiple locality descriptions as sources of spatial evidence to infer its location. 

For example, the locality description \emph{“Main Divide, Coromandel Ra, 1 mile SE of Maumaupaki Trig”} indicates that a specimen was collected one mile southeast of Maumaupaki Trig, a trig station in the Maumaupaki area of New Zealand that is absent from current gazetteers. Because the specimen record includes geographic coordinates (latitude: $-36.97612$, longitude: $175.5862$), this relation can be inverted to infer that the NGP lies one mile northwest of the specimen location.

When multiple descriptions refer to the same NGP using different spatial relations, as illustrated in Table~\ref{tab:cabstand-example}, the intersection of these constraints enables a more precise estimate of its location. The following subsections describe how this strategy is operationalised in our study.

\subsubsection{Interpreting spatial relations}
\label{sec:interpreting-spatial-relations}

Spatial relations in specimen locality descriptions are represented as triplets comprising a relatum (the referenced place), a spatial relation, and a locatum (the target location). For example, in the locality description “\emph{18 km west of Ashburton, on Maronan Valetta Road}”, \emph{Ashburton} is the relatum, \emph{18 km west of} specifies the spatial relation, and the locatum is the specimen’s collection location, which is implicit in the text and represented by the associated coordinates found in the specimen record. 

Assuming that the place name to be georeferenced is \emph{Ashburton}, the spatial relation can be inverted, treating the known specimen location as the relatum and the place name as the locatum. Locality descriptions may include additional place names; similar to the example above, “on Maronan Valetta Road” indicates that the specimen was collected on Maronan Valetta Road and does not express a direct spatial relationship between Maronan Valetta Road and Ashburton. Accordingly, our methodology considers only direct spatial relationships between the target place name and the specimen location.

We also acknowledge an asymmetry between relatum and locatum, as described in Talmy’s work \cite{talmy2000toward}, which highlights differences between the reference object and the object whose location is being described. Reference objects are typically larger, more permanently located, more geometrically complex, and more independent than the objects they help locate. For example, \emph{“the post box is by the post office”} is a more natural expression than \emph{“the post office is by the post box”} because the post office is larger and more permanent, making it a more plausible reference object. However, this asymmetry does not pose a problem for our georeferencing methods, as in our data the locatum is a fixed point observation (the specimen location), and multiple such observations are aggregated to derive spatial constraints.

\begin{figure}
    \centering
    \includegraphics[width=0.85\linewidth]{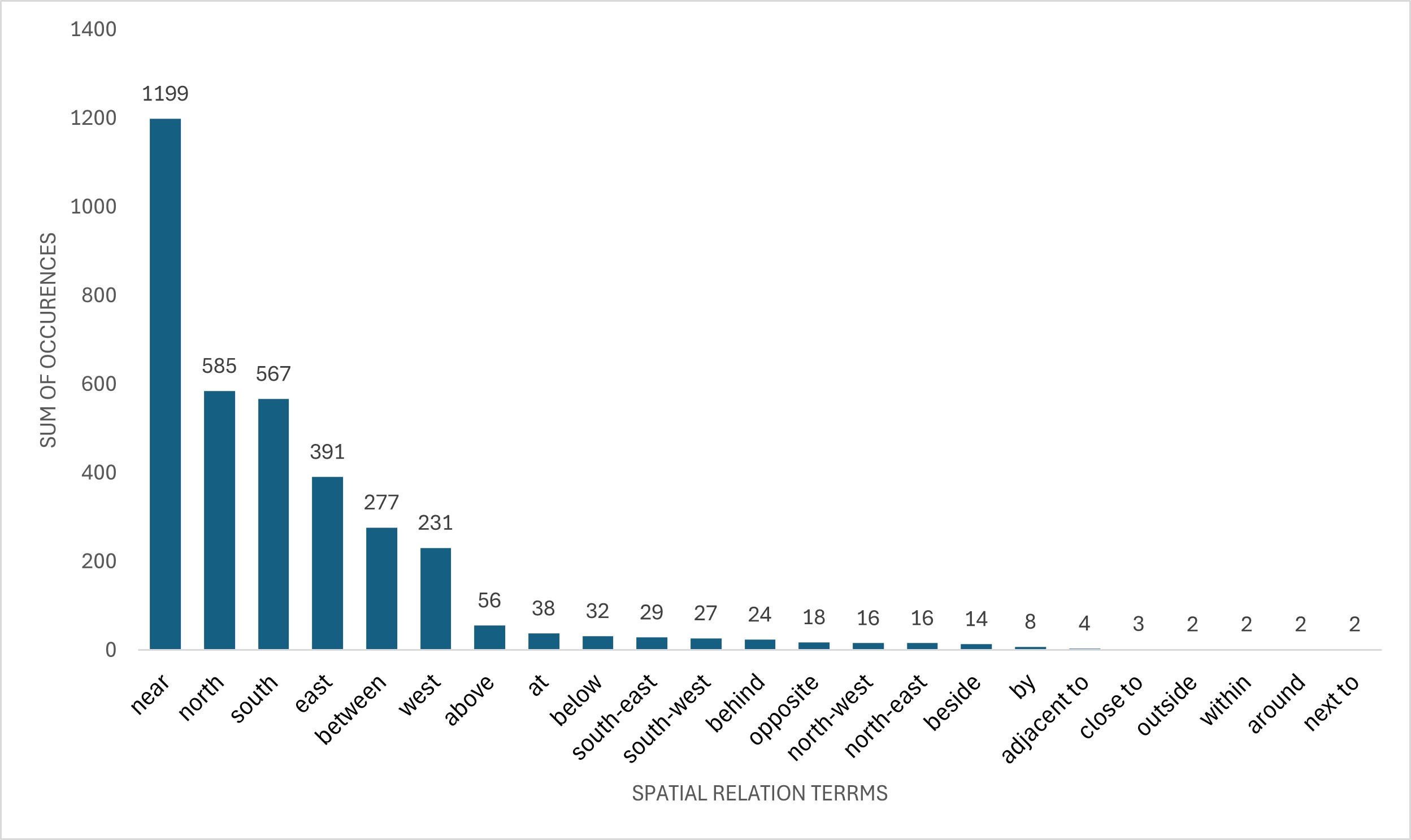}
    \caption{Frequency distribution of spatial relation terms extracted from specimen locality descriptions.}
    \label{fig:spatial-terms-all}
\end{figure}

In our dataset, we observed a wide range of spatial relation terms (Figure~\ref{fig:spatial-terms-all}). The distribution is highly skewed: the proximity relation \emph{near} occurs substantially more frequently than any other term. Cardinal direction relations (\emph{north}, \emph{south}, \emph{east}, and \emph{west}) also appear frequently, whereas the remaining terms occur far less often. In addition, cardinal and inter-cardinal relations (e.g., south-west, north-east) are sometimes accompanied by an explicit distance, e.g., \emph{``2~km north of''}, which strengthens the spatial constraint beyond direction alone.

Examining the relation terms in our dataset, we attempted to categorise them. 
However, a key challenge in adopting categorisations from prior studies is that our use case involves inverting the locatum and relatum implied by the locality description; consequently, any categorisation must remain meaningfully interpretable under inversion. While the inverses of cardinal and inter-cardinal relation terms are straightforward, many other relations appear to be symmetrical, such that their interpretation remains unchanged or similar when inverted.

For these non-directional expressions, we therefore treat them as proximity-like constraints even though the perceived degree of proximity could vary. Terms such as \emph{near}, \emph{by}, \emph{close to}, \emph{next to}, \emph{adjacent to}, and \emph{beside} are clearly proximity-related and effectively symmetric (e.g., if $X$ is near $Y$, then $Y$ is near $X$). Containment-oriented expressions such as \emph{at}, \emph{within}, and \emph{outside} instead constrain the specimen location relative to a region associated with the referenced feature. However, when the relation is inverted, they do not specify a unique direction or operator for locating the target place name; operationally, they still primarily imply geographic closeness to the relatum. Because specimen records represent the specimen location as a point (latitude/longitude), statements such as ``the specimen is within/outside $X$'' ($X$ is a place name) are best interpreted as indicating proximity to $X$, with the direction left unspecified. Orientation-based terms such as \emph{above}, \emph{below}, \emph{behind}, and \emph{opposite} often depend on a local frame of reference, and thus are not consistently invertible from text alone. Across these cases, modelling non-cardinal relations as proximity-like constraints appears to preserve their role as locality restrictions while avoiding misinterpretation in the inverted setting.

\subsubsection{Evaluation dataset design and representation}

\paragraph*{Pseudo-NGP dataset for evaluation}

To evaluate our methods for georeferencing, we constructed a subset from the preprocessed dataset of 13,746 locality descriptions mentioned in Section \ref{sec:data-prep}. To enable evaluation, we required place names with known coordinates as ground-truth data. Therefore, we selected locality descriptions that contain spatial relations referring to place names recorded in the NZGB Gazetteer. Place names were disambiguated and resolved against the NZGB Gazetteer using the \emph{state/province} information provided in the corresponding biological specimen records. As this study serves as a proof of concept, only place names represented as point geometries were selected in order to simplify the evaluation setting. These gazetteered place names were treated as pseudo-NGPs by withholding their coordinates during inference, allowing us to quantitatively assess the accuracy of the proposed methods. This dataset thus provides a benchmark for systematic evaluation.

Among the resolved place names, we further selected those referenced in at least three locality descriptions, as three constraints constitute the minimum required to define a bounded area. Under this criterion, the final evaluation dataset consists of 2,403 locality descriptions referring to 365 place names. Each locality description is associated with specimen coordinates, and each place name has corresponding resolved ground-truth coordinates.

\paragraph*{Input data representation}

Both proposed deterministic and probabilistic methods operate over a structured JSON representation derived from the dataset. For each target place name, we extract spatial relation mentions from locality descriptions and group them into a single place-centric entry. Each entry follows the schema  \emph{place\_name} $\rightarrow$ \emph{place\_name\_coordinates} $\rightarrow$ \emph{relations}:

\begin{itemize}
    \item \textbf{place\_name}: identifier for the target place.
    \item \textbf{place\_name\_coordinates}: ground-truth latitude/longitude of the place\_name (retained only for evaluation and not used to form constraints).
    \item \textbf{relations}: a list of spatial relations associated with the \emph{place\_name}, extracted from specimen locality descriptions as described in Section \ref{sec:data-prep}. Each relation entity includes the extracted spatial relation term, an associated distance value when provided, and the coordinates of the specimen record from which the relation was extracted. Relation terms are normalised to a canonical form to reduce surface variation (e.g., collapsing whitespace, standardising hyphenation for multiword expressions such as \emph{next to} $\rightarrow$ \emph{next-to}, and mapping direction variants such as \emph{NE}, \emph{north east} to a single label). Cardinal and inter-cardinal relations are represented in their inverted form, consistent with our interpretation that the specimen coordinate is the relatum and the target place name is the locatum.
\end{itemize}

We have used the place name \emph{Kauangaroa}\footnote{\emph{Kauangaroa} is a locality in Wellington, New Zealand.} as the running example to illustrate this process. Table~\ref{tab:kauangaroa-spatial-relations} lists three locality descriptions that reference \emph{Kauangaroa}, along with the specimen coordinates associated with each description and the extracted (and, where applicable, inverted) spatial relation terms. Listing~\ref{lst:place-graph-json-kauangaroa} shows the corresponding JSON entry produced from the data in Table~\ref{tab:kauangaroa-spatial-relations}. We generated one such JSON entry per place name and combined them into a single JSON file covering all 365 places in our dataset as the input.

\begin{lstlisting}[caption={Example JSON entry used to generate the ALR for the target place name \emph{Kauangaroa}},label={lst:place-graph-json-kauangaroa}, basicstyle=\ttfamily\footnotesize]
{
  "place_data": [{
      "place_name": "kauangaroa",
      "place_name_coordinates": {
        "latitude": "-39.928234", "longitude": "175.279616"
        },
      "relations": [
        {
          "relation": "near",
          "distance": "",
          "latitude": "-39.93045", "longitude": "175.2665"
        },
        {
          "relation": "west",
          "distance": "1km",
          "latitude": "-39.92546", "longitude": "175.2909"
        },
        {
          "relation": "east",
          "distance": "",
          "latitude": "-39.91161", "longitude": "175.2635"
        }
    }]
}
\end{lstlisting}

\begin{table}[b]
\caption{Locality descriptions from specimen records that refer to \emph{Kauangaroa} using relative spatial expressions. The latitude and longitude values correspond to the coordinates of the specimen locations described in each locality description.}{
\centering
\footnotesize
\setlength{\tabcolsep}{4pt}
\begin{tabularx}{\textwidth}{@{}l >{\raggedright\arraybackslash}X r r >{\raggedright\arraybackslash}p{2.4cm} >{\raggedright\arraybackslash}p{2.4cm}@{}}
\toprule
\textbf{Place Name} & \textbf{Locality Description} & \textbf{Latitude} & \textbf{Longitude} & \textbf{Extracted Spatial Relation} & \textbf{Inversed Spatial Relation} \\
\midrule
\multirow{3}{*}{Kauangaroa}
  & Near Fordell at Kauangaroa
  & -39.93045 & 175.2665
  & near
  & near (R1) \\ \midrule
& 1 km east of Kauangaroa, Whangaehu River
  & -39.92546 & 175.2909
  & 1 km east
  & 1 km west (R2) \\ \midrule
& Whangaehu River catchment west of Kauangaroa
  & -39.91161 & 175.2635
  & west
  & east (R3) \\
\bottomrule
\end{tabularx}}
\label{tab:kauangaroa-spatial-relations}
\end{table}

Building on this representation, we now describe the proposed methods for georeferencing NGPs.

\subsubsection{Deterministic georeferencing using spatial relations}
In this approach, we framed the georeferencing of NGPs as an inference problem over a \emph{place graph}. The \emph{place graph} provided a compact representation that connects each target place name to all specimen locations whose locality descriptions mention that place name. We then aggregated the resulting spatial constraints to derive an \emph{Approximate Location Region} (ALR) for each target place $p$. The key steps of this method are as follows: 

\begin{itemize}
\item \textbf{Relatum point set:} For each relation entity in the JSON file generated, we read the specimen coordinates (latitude/longitude) and represented them as relatum points. These points served as the geometric anchors from which constraint regions were generated.

\item \textbf{Distance parsing:} When a distance value was present, it was parsed and converted to kilometres (from meters or miles). If no distance was provided, it was treated as unspecified and handled according to the relation-specific defaults described below.

\item \textbf{Spatial context region:} For each target place $p$, we derived a bounded spatial context to constrain subsequent geometric operations. This context was defined as the bounding box of the relatum point set, expanded by a buffer determined from the median nearest-neighbour distance among the relatum points. This statistic also provided an estimate of the local spread of observations associated with $p$.

\item \textbf{Constraint region construction:} Each relation entity was converted into a feasible region within the spatial context:
\begin{itemize}
    \item For cardinal and inter-cardinal relations, we generated a directional region by intersecting the spatial context with the corresponding half-plane(s). If a distance was given, the region was additionally intersected with a distance buffer around the relatum point.
    \item For all remaining relations, we generated a proximity-like buffer around the relatum point. When an explicit distance was not provided, we set the buffer radius adaptively based on how tightly the specimen (relatum) points for that place name cluster, using their median nearest-neighbour spacing as a proxy for local spread. The resulting default radius was then clipped to global bounds of 0.2~km and 30~km. The lower bound of 0.2 km prevents unrealistically small buffers that would frequently produce empty intersections given coordinate uncertainty and vague locality descriptions, while the upper bound of 30 km prevents weak proximity relations from dominating the spatial context with overly broad regions.
\end{itemize}

\item \textbf{ALR computation:} The ALR for $p$ is computed as the intersection of all retained constraint regions. 

\end{itemize}

\begin{figure*}[h]
    \centering
    \begin{subfigure}[t]{0.495\textwidth}
        \centering
        \includegraphics[width=\linewidth]{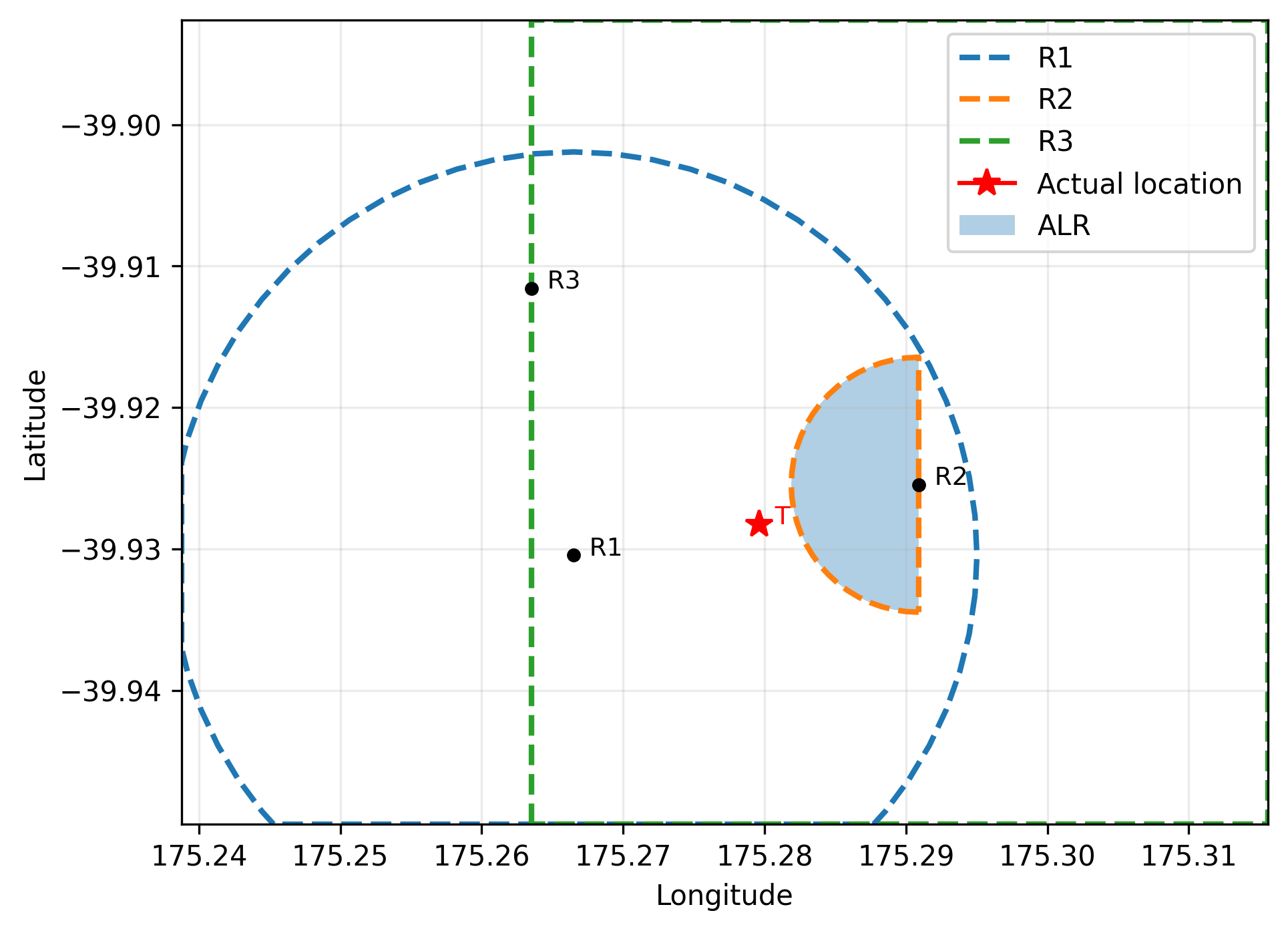}
        \caption{Constrain regions and ALR for \emph{Kauangaroa}.}
        \label{fig:place-graph-kauangaroa}
    \end{subfigure}
    \hfill
    \begin{subfigure}[t]{0.495\textwidth}
        \centering
        \includegraphics[width=\linewidth]{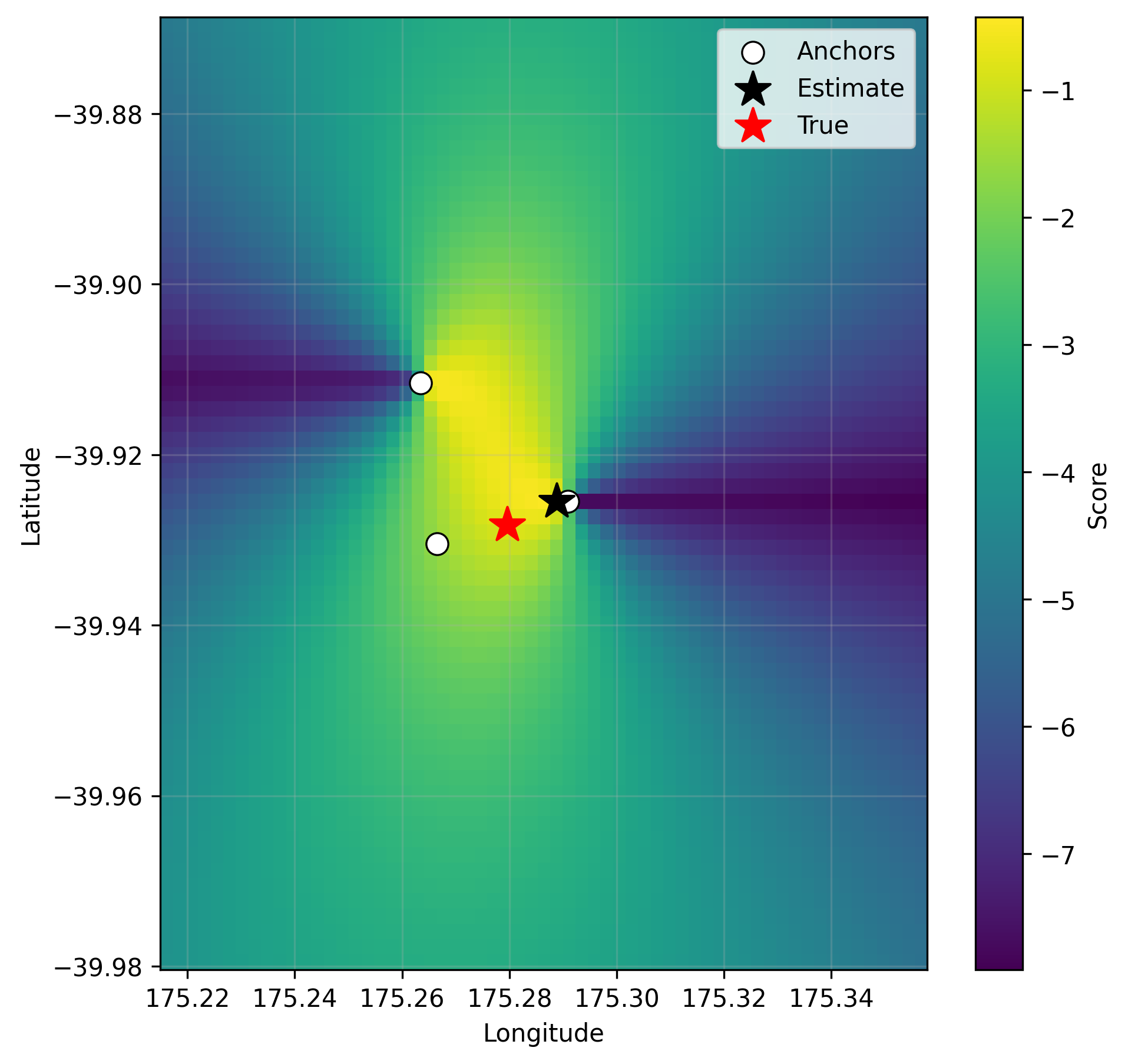}
        \caption{Probabilistic score surface for \emph{Kauangaroa}.}
        \label{fig:heatmap-kauangaroa}
    \end{subfigure}
    
    \caption{Deterministic (a) and probabilistic (b) georeferencing results for \emph{Kauangaroa}. The prediction errors between the actual and estimated locations are 0.71 km and 0.83 km, respectively.}
    \label{fig:kauangaroa-comparison}
\end{figure*}

Figure~\ref{fig:place-graph-kauangaroa} illustrates the spatial constraint regions generated for \emph{Kauangaroa} from the JSON schema in Listing~\ref{lst:place-graph-json-kauangaroa}. The figure shows the feasible region associated with each relation (ordered as in the schema) and the resulting ALR obtained as the intersection of these constraint regions. In this formulation, the ALR provides a feasible region rather than a uniquely determined coordinate; accordingly, for evaluation, we used the centroid of the ALR as a point estimate and interpreted the ALR as an uncertainty region. However, an ALR was not obtained for all places: in some cases, the strict intersection became empty due to incompatible or overly restrictive constraints across locality descriptions, particularly when proximity-like relations were imprecise or when extracted relations contain noise. This limitation motivated our second approach, which replaced hard region intersection with a probabilistic formulation to produce a model-derived point estimate.

\subsubsection{Probabilistic georeferencing using spatial relations}

In this method, we replaced region intersection with a likelihood-based formulation that evaluated candidate coordinates against all extracted spatial relations for a given place name. Each relation contributed a likelihood (soft preference) around its specimen anchor, and the location that maximised the aggregated likelihood was selected as the final coordinate estimate. The key steps of this method are as follows: 

\begin{itemize}
\item \textbf{Relatum point set:} As before, each relation entity provided a specimen coordinate that was treated as the relatum point anchoring the constraint. Distances, when present, were parsed and converted to kilometres.

\item \textbf{Candidate search space:} For each place name $p$, we defined a bounded candidate region based on the specimen locations that reference $p$.
\begin{itemize}
    \item We computed a robust spatial centre as the median latitude and median longitude of all referencing specimen coordinates.
    \item We defined a fixed-radius spatial bound (60~km) around this centre to limit the candidate region and reduce unlikely solutions far from the evidence. This radius is approximately four times the estimated proximity uncertainty parameter ($s_{\text{near}}$) described below, beyond which likelihood contributions become negligible. This fixed-radius search assumes that the specimen anchors for a target place form a spatially coherent set. It may be less appropriate for records containing strong outliers or multiple spatial clusters, where an adaptive or cluster-aware search region would be preferable.
\end{itemize}
\item \textbf{Likelihood construction:} For each relation, we define a log-likelihood $\ell(x)$ at candidate location $x$ relative to anchor $a$, where $d(x,a)$ denotes the great-circle (haversine) distance and $\Delta\theta(x,a)$ the angular deviation from the expected bearing; $d_0$ is the stated distance when provided. Proximity is modelled via distance decay, while directional and distance deviations are modelled using Gaussian (normal) distributions.
\begin{itemize}
    \item \emph{Proximity:} 
    $\ell_{\text{prox}}(x) = - d(x,a) / s_{\text{near}}$.

    \item \emph{Directional:} 
    $\ell_{\text{dir}}(x) = - \Delta\theta(x,a)^2 / (2\sigma_{\theta}^2)$.

    \item \emph{Directional + distance:} 
    $\ell_{\text{dir+dist}}(x) =
    - \Delta\theta(x,a)^2 / (2\sigma_{\theta}^2)
    - (d(x,a)-d_0)^2 / (2\sigma_d^2)$.
\end{itemize}

Here $s_{\text{near}}$ (km) controls proximity decay, 
$\sigma_{\theta}$ (degrees) directional uncertainty, 
and $\sigma_d$ (km) distance uncertainty.



\item \textbf{Uncertainty parameter estimation:} As described above, the log-likelihood construction depended on three uncertainty parameters: $s_{\text{near}}$, $\sigma_{\theta}$, and $\sigma_{d}$. To estimate these values, we used an auxiliary subset of our preprocessed dataset consisting of place names that appeared in only one or two locality descriptions. These place names are listed in the NZGB Gazetteer but were excluded from the evaluation dataset, which requires at least three descriptions per place name. The auxiliary subset contained 1,485 locality descriptions referring to 1,204 resolved place names. Using the ground-truth coordinates of these place names, we computed: (i) anchor-to-ground-truth distances for proximity-like relations, (ii) angular deviations between observed bearings and expected directions for directional relations, and (iii) differences between observed and stated distances for distance-qualified relations. Because the distance-related distributions included extreme values, we used the 90th percentile of the distance and absolute distance-difference distributions to determine $s_{\text{near}}$ and $\sigma_{d}$, thereby representing the majority of cases. For directional relations, we used the median angular deviation to determine $\sigma_{\theta}$. The resulting parameter values were $s_{\text{near}} = 14$\,km, $\sigma_{\theta} = 47^\circ$, and $\sigma_{d} = 4$\,km.

\item \textbf{Combining evidence:} For a given candidate location, log-likelihoods were summed across all relations to obtain a single score.

\item \textbf{Coarse-to-fine MAP search:} The bounded region was discretised into an initial coarse grid with 3~km spacing, enabling an efficient first-pass search while preserving sufficient spatial coverage. Each grid point was evaluated by summing relation log-likelihoods, and the best-scoring point was retained. We then iteratively refined the estimate by defining a smaller square window centred on the current best point. The window extent was determined relative to the previous grid spacing, covering a local neighbourhood of candidate cells defined as a square region spanning a fixed multiple of the current grid resolution in each direction. Within this window, progressively finer grid spacings (1.2~km, 0.5~km, and 0.2~km) were applied. At each level, candidates were re-scored, and the maximum was retained. The final latitude/longitude estimate was the highest-scoring point at the finest resolution.

\end{itemize}

Overall, the method produces a single coordinate for each target place name by selecting the candidate that maximises the aggregated likelihood across all extracted relations. Figure~\ref{fig:heatmap-kauangaroa} visualises the resulting probabilistic surface for \emph{Kauangaroa}. Warmer colours indicate candidate locations that are more consistent with the extracted spatial relations when evaluated jointly against all specimen anchor points (white markers), and the peak of this surface (black star) corresponds to the model’s preferred coordinate estimate.

\subsubsection{LLM-based georeferencing}
As the third approach, we explored how an LLM performs on this spatial modelling style georeferencing task, using the \texttt{gpt-5.1-2025-11-13} model via the OpenAI Batch API\footnote{https://developers.openai.com/api/docs/guides/batch/} with a simple zero-shot prompt. Figure~\ref{fig:gpt-prompt}a presents the inference prompt; only the locality descriptions and coordinates differ across target places. No system prompt was used, and we did not modify the default model settings.

Because our pseudo-NGP dataset consists of real gazetteered toponyms with coordinates withheld for evaluation, we replaced each target name in the prompt with a placeholder (\emph{Kolombo}) to prevent the model from relying on prior knowledge of the name. The model was explicitly instructed to use only the provided locality descriptions and coordinates as evidence, allowing us to evaluate the ability of a general-purpose LLM to perform spatial modelling from textual spatial information. In a true-NGP setting, however, an LLM may still draw on latent knowledge from historical documents, web pages, maps, or other textual sources seen during training. Such knowledge could improve georeferencing when accurate, but it may also introduce unsupported assumptions. Thus, masking provides a stricter evaluation of inference from specimen-based evidence alone, while true-NGP georeferencing would require further evaluation of how LLMs combine provided evidence with their inherent geographic knowledge.

The LLM was prompted to return a single coordinate estimate for each target place name using the provided locality descriptions and specimen coordinates. To improve transparency, the model was also asked to return its reasoning alongside the predicted coordinates. We found that including this reasoning step also improved the overall accuracy of the LLM-based method, as reported in Table~\ref{tab:georef-results}. Figure~\ref{fig:gpt-prompt}b presents the verbatim response from the LLM, including the reasoning behind the inferred coordinates.

\begin{figure}[h]
\caption{(a) Prompt used to georeference \emph{Kauangaroa} with GPT-5.1. (b) Verbatim response showing the detailed reasoning. The prediction error is 0.8 km.}
\label{fig:gpt-prompt}

\setlength{\fboxsep}{6pt}      
\setlength{\fboxrule}{0.5pt}   

\fbox{%
\begin{minipage}{0.97\columnwidth}

\begin{quote}\footnotesize
\textbf{(a) Prompt:}

\textbf{Task:} Determine the coordinates of `Kolombo'.

\textbf{Constraints:}
\begin{itemize}\setlength{\itemsep}{0pt}
\item Use \textbf{ONLY} the locality descriptions and coordinates provided below as evidence.
\item Do \textbf{NOT} use prior knowledge, assumptions, or any external sources.
\item If multiple locations are described, infer Kolombo’s position based on their spatial relationships.
\item If the evidence is insufficient or contradictory, state that explicitly.
\end{itemize}

\textbf{Evidence (multiple locality descriptions):}

\textbf{1)} Near Fordell at Kolombo. 
Lat = -39.93045, Lon = 175.2665

\textbf{2)} 1 km east of Kolombo, Whangaehu River. 
Lat = -39.92546, Lon = 175.2909

\textbf{3)} Whangaehu River catchment west of Kolombo. 
Lat = -39.91161, Lon = 175.2635
\end{quote}

\hrule

\vspace{0.5em}

\begin{quote}\footnotesize
\textbf{(b) Response (Verbatim):}

We have three points described relative to Kolombo:

\textbf{1)} “Near Fordell at Kolombo”: (-39.93045, 175.2665). This is essentially at Kolombo or very close to it.

\textbf{2)} “1 km east of Kolombo, Whangaehu River”: (-39.92546, 175.2909). If this point is east of Kolombo, then Kolombo must lie to its west, i.e. at a smaller longitude than 175.2909.

\textbf{3)} “Whangaehu River catchment west of Kolombo”: (-39.91161, 175.2635). If this point is west of Kolombo, then Kolombo must lie to its east, i.e. at a larger longitude than 175.2635.

Combining (2) and (3), Kolombo’s longitude must lie between 175.2635 and 175.2909. The point explicitly described as “Near Fordell at Kolombo” at 175.2665 falls within this interval, making it a strong candidate.

For latitude, the “at Kolombo” point is -39.93045. The other two sites lie slightly north, consistent with them being nearby but distinct locations.

Thus, the best estimate for Kolombo is approximately -39.930, 175.267 (rounded slightly for representation).
\end{quote}

\end{minipage}%
}

\end{figure}

\section{Results and Discussion}
\label{sec:results}

To evaluate performance, we report the mean and median error distance (km) between predicted and true coordinates. We also compute A@161 and A@1, measuring the percentage of place names correctly georeferenced within 161 km (100 miles) and 1 km, respectively. A@161 is widely used in toponym resolution and geocoding \cite{ DeLozier_Baldridge_London_2015, gritta2018s, sharma2023spatially} to capture coarse regional accuracy, while A@1 provides a stricter measure of spatial accuracy.

Table~\ref{tab:georef-results} compares the georeferencing performance of the deterministic, probabilistic, and LLM-based methods across these metrics for predicting coordinate estimates for all 365 pseudo-NGP place names in the evaluation dataset. The deterministic model shows the weakest overall performance. It fails to produce estimates for 108 place names, reflecting its reliance on strict rule satisfaction and its inability to handle ambiguous or conflicting spatial constraints. Even when estimates are produced, it records the highest mean error (9.19 km) and the lowest A@1 and A@161 scores.

Both the probabilistic and LLM-based methods successfully generate location estimates for all place names and achieve identical A@161 values, with only two shared outliers exceeding the 161 km threshold. Between these two approaches, the probabilistic model performs best overall, achieving the lowest median error (1.43 km) and the highest A@1 score, indicating more precise fine-grained georeferencing. The two GPT-5.1 prompt variants used for the LLM-based method show broadly comparable performance. Overall, the reasoning prompt performs slightly better in terms of distance-based error, reducing both the mean and the median errors, although the direct prompt achieves a higher A@1 score. The LLM-based variants achieve the lowest mean errors overall, suggesting fewer extreme failures; however, their higher median errors and lower A@1 scores indicate reduced accuracy at finer spatial scales compared to the probabilistic approach.

We examined the two place names for which both the probabilistic and LLM-based methods produced errors exceeding 161 km: Fog Peak and Rockdale. In both cases, the specimen coordinates lie far from the corresponding NZGB gazetteered locations. Fog Peak is an off-track mountain peak, which may contribute to ambiguity in locality descriptions, while the gazetteered location of Rockdale is distant from the specimen coordinates, despite descriptions stating the specimens are “near Rockdale”. These discrepancies likely reflect issues in the original data rather than failures of the georeferencing methods. 

\begin{table}[ht]
\centering
\small
\caption{Comparison of the georeferencing performance of the three methods. Mean and median denote error distance in kilometres. ↓ indicates lower is better; ↑ indicates higher is better.}
\label{tab:georef-results}
\begin{tabularx}{\textwidth}{l *{5}{>{\centering\arraybackslash}X}}
\hline
Model & Mean error↓ & Median error↓ & A@1↑ & A@161↑ & No estimate↓ \\
\hline
Deterministic model & 9.19 & 1.72 & 20.3\% & 69.6\% & 108 \\
Probabilistic model & 8.95 & \textbf{1.43} & \textbf{36.0\%} & \textbf{99.5\%} & 0 \\
GPT-5.1 & 6.23 & 1.85 & 30.9\% & \textbf{99.5\%} & 0 \\
GPT-5.1 with reasoning & \textbf{5.89} & 1.80 & 28.1\% & \textbf{99.5\%} & 0 \\

\hline
\end{tabularx}
\end{table}

\subsection{Performance based on the spatial relations}

To further investigate the results of the three models, we analysed model performance with respect to the spatial relations present in locality descriptions associated with each place name. Spatial relation terms were categorised into three groups: direction, distance with direction, and proximity, which provided additional insight into the types of relations each model handled more effectively. Table~\ref{tab:spatial-relations-metrics} summarises model performance across different combinations of spatial relations. In this table, the LLM-based method refers to the GPT-5.1 reasoning-prompt setting.

The results indicate that although the probabilistic model performs well overall, it is less effective when only directional relations are available, underperforming both the deterministic and LLM-based approaches in this setting. In contrast, when only proximity relations are present, the probabilistic model performs comparatively well. When locality descriptions include all three types of spatial constraints, direction, distance, and proximity, both the probabilistic and LLM-based methods achieve strong performance, with the probabilistic model yielding the highest overall accuracy. Notably, the LLM-based method exhibits relatively consistent performance across different types of relational constraints.

These differences can be partly explained by how each method interprets and integrates different types of spatial relations. Direction-only relations constrain bearing but not distance, leaving the location of the likelihood peak along the directional axis underdetermined. As a result, the probability surface remains diffuse, limiting spatial precision in the absence of additional constraints. Proximity relations, by contrast, align well with probabilistic representations of spatial uncertainty, allowing nearby candidate locations to be weighted without precise orientation. Explicit distance constraints act as strong anchors for all methods, particularly when combined with directional information.

\begin{table}[t]
\centering
\small
\caption{Comparison of georeferencing performance of the three methods by spatial relation type}
\label{tab:spatial-relations-metrics}
\setlength{\tabcolsep}{4pt}
\renewcommand{\arraystretch}{1.05}

\begin{tabular}{l l r r}
\toprule
\textbf{Spatial relations found} &
\textbf{Model} &
\textbf{\shortstack{Mean error\\(km)↓}} &
\textbf{\shortstack{Median error\\(km)↓}} \\
\midrule

Direction only & Deterministic model & 10.02 & 2.30 \\
Direction only & Probabilistic model & 34.58 & 5.25 \\
Direction only & LLM-based model & \textbf{6.46} & \textbf{1.41} \\
\midrule

Proximity only & Deterministic model & 6.87 & 1.56 \\
Proximity only & Probabilistic model & \textbf{7.33} & \textbf{1.09} \\
Proximity only & LLM-based model & 7.48 & 1.32 \\
\midrule

Direction + Distance & Deterministic model & 4.02 & 2.39 \\
Direction + Distance & Probabilistic model & 7.60 & \textbf{0.74} \\
Direction + Distance & LLM-based model & \textbf{2.59} & 2.02 \\
\midrule

Direction + Proximity & Deterministic model & 35.01 & 2.25 \\
Direction + Proximity & Probabilistic model & 8.85 & \textbf{1.58} \\
Direction + Proximity & LLM-based model & \textbf{7.98} & 1.94 \\
\midrule

Direction + Distance + Proximity & Deterministic model & 33.71 & 1.57 \\
Direction + Distance + Proximity & Probabilistic model & \textbf{2.73} & \textbf{1.47} \\
Direction + Distance + Proximity & LLM-based model & 2.81 & 1.82 \\
\bottomrule
\end{tabular}
\end{table}

\subsection{Key observations on LLM reasoning}
We examined the reasoning returned by GPT-5.1 to understand the behaviours that enabled it to georeference effectively across the dataset. Focusing on cases where the LLM outperformed the probabilistic and deterministic methods, two patterns stand out.

First, as shown in Figure~\ref{fig:gpt-prompt}b, GPT-5.1 is effective at extracting spatial information directly from free text and synthesising it into a coherent interpretation without manual preprocessing. It correctly identifies spatial relations in the text and incorporates associated distance information without relying on an explicit parser. Notably, it independently adopted our approach: inverting the spatial relation so that it is interpreted from the specimen location toward the referenced place name. We also observe that GPT-5.1 accurately performs geographic calculations by combining distance and direction cues, for example, converting stated offsets into rough $\Delta\mathrm{lat}/\Delta\mathrm{lon}$ adjustments and performing basic unit conversions (e.g., miles to kilometres).

Second, the model does not treat each reference equally. Its inference strategy is to find a best-fit solution while ignoring noise, instead of trying to satisfy every relational mention. Rather than weighting all relations equally, the LLM emphasises a subset of mutually consistent (or otherwise more reliable) constraints when deriving an estimate. This behaviour can be observed in the reasoning shown in Figure~\ref{fig:gpt-prompt}b, in which the LLM primarily uses the “at” coordinates to infer the final coordinates, while the other mentions are used mainly to validate plausibility. This substantially benefits the LLM approach, contributing to the lowest mean error.

\subsection{Summary of results}
Table \ref{tab:method-pros-cons} provides a summary of the strengths and limitations of each model.

\begin{table}[h]
\centering
\small
\caption{Qualitative comparison of georeferencing approaches based on observed behaviour in the experiments.}
\label{tab:method-pros-cons}
\renewcommand{\arraystretch}{1.1}
\setlength{\tabcolsep}{4pt}

\begin{tabularx}{\linewidth}{@{} 
    p{2.3cm} 
    >{\raggedright\arraybackslash}X 
    >{\raggedright\arraybackslash}X 
@{}}
\toprule
\textbf{Method} & \textbf{Strengths} & \textbf{Limitations} \\
\midrule

Deterministic &
Simple and interpretable; Explicitly models an uncertainty region using ALR; Effective when inputs are precise; Exact constraint satisfaction ensures reproducibility. &
Requires structured inputs and explicit relation modelling; Sensitive to minor inconsistencies, resulting in the lowest performance. \\
\midrule
Probabilistic &
Highest overall precision, with the best median error and Acc@1; Performs well when proximity constraints (the most common) are present; Generates estimates for all cases; Reproducible under fixed parameters. &
Requires structured inputs and explicit relation modelling; Performs less well with direction-only relations; Affected by inconsistent or imprecise relations. \\
\midrule
LLM-based &
Operates directly on unprocessed locality descriptions; Produces acceptable estimates for most cases (lowest mean error); Flexible in handling a wide variety of spatial relations (not limited to a pre-defined set). &
Reproducibility is not guaranteed; Less precise than the probabilistic model; Costly. \\
\bottomrule
\end{tabularx}
\end{table}


\section{Conclusion}
\label{sec:conclusion}
This study shows that biological specimen records are a practical source for both identifying and georeferencing non-gazetteered place names within a region. We establish the presence of NGPs in specimen collection records and provide an empirical estimate of their prevalence in the collection. We also compare traditional deterministic and probabilistic georeferencing approaches with a modern LLM-based approach to assess whether LLMs can reliably support a largely spatial modelling–based task. The results indicate that, while the LLM-based method yields broadly acceptable estimates, traditional spatial modelling, particularly probabilistic inference, remains more precise when fine-grained localisation is required.

\textbf{Limitations.} The proposed methods are currently implemented only for NGPs that can be represented as points, limiting applicability to linear or areal features. Performance depends on the availability, spatial distribution, and accuracy of georeferenced specimen anchors, as well as on the diversity of spatial relations available for each place name. The performance of the methods may therefore be biased toward better-constrained place names. In addition, the deterministic and probabilistic methods rely on predefined spatial relation terms and predefined interpretations of those relations. These choices may not fully capture the variability, semantic differences, and context dependence of spatial language. Place name recognition and relation extraction from locality descriptions also remain challenging for deterministic and probabilistic methods due to vague phrasing and inconsistent recording practices, and would benefit from more robust extraction methods.

\textbf{Future work.} Future extensions will support non-point geometries (lines and polygons) and richer representations of uncertainty. We also aim to improve spatial relation parsing to capture both explicit and implicit relations. The LLM-based method could also be further improved through more systematic prompt design, including in-context learning with few-shot examples, and exploration of uncertainty estimation from model outputs. We further aim to develop a hybrid pipeline that combines the strengths of LLMs with probabilistic inference to georeference NGPs. Finally, we will expand the analysis to additional regions, particularly those with limited historical coverage in existing gazetteers, alongside improved methods for NGP recognition.

\bibliography{bib2doi}

\end{document}